\documentclass[sigconf,nonacm]{acmart}
\AtBeginDocument{%
  }

\usepackage{graphicx}       
\usepackage{array}           
\usepackage{booktabs} 
\usepackage{colortbl} 
\usepackage{multirow}        
\usepackage{hhline}          
\usepackage{caption}        
\usepackage{tabularx} 
\usepackage{booktabs}
\usepackage{makecell}
\usepackage{enumitem}
\usepackage{tcolorbox}
\usepackage{booktabs}
\usepackage{adjustbox}
\usepackage{fancyhdr}
\usepackage{fontawesome}

\begin{document}

\title[Embodied-R: Collaborative Framework for Activating Embodied Spatial Reasoning in Foundation Models via Reinforcement Learning]{Embodied-R: Collaborative Framework for Activating\\ Embodied Spatial Reasoning in Foundation Models via \\ Reinforcement Learning}

\author{
    \textbf{Baining Zhao}$^{*}$,
    \textbf{Ziyou Wang}$^{*}$,
    \textbf{Jianjie Fang}$^{*}$,
    \textbf{Chen Gao}$^{\dagger}$,
    \textbf{Fanghang Man},\\
    \textbf{Jinqiang Cui},
    \textbf{Xin Wang}, 
    \textbf{Xinlei Chen}$^{\dagger}$,
    \textbf{Yong Li},
    \textbf{Wenwu Zhu}\\
    Tsinghua University\\ 
    \textcolor{magenta}{\faGlobe\ \href{https://embodiedcity.github.io/Embodied-R/}{\textcolor{magenta}{\textit{Project Page}}}} \hspace{3cm}
    \textcolor{black}{\faGithub\ \href{https://github.com/EmbodiedCity/Embodied-R.code}{\textcolor{black}{\textit{Code}}}}
}

\pagestyle{fancy}
\fancyhf{}
\fancyhead[L]{} 
\fancyhead[R]{}
\fancyfoot[C]{} 

\begin{abstract}
Humans can perceive and reason about spatial relationships from sequential visual observations, such as egocentric video streams. However, how pretrained models acquire such abilities, especially high-level reasoning, remains unclear. This paper introduces Embodied-R, a collaborative framework combining large-scale Vision-Language Models (VLMs) for perception and small-scale Language Models (LMs) for reasoning. Using Reinforcement Learning (RL) with a novel reward system considering think-answer logical consistency, the model achieves slow-thinking capabilities with limited computational resources. After training on only 5k embodied video samples, Embodied-R with a 3B LM matches state-of-the-art multimodal reasoning models (OpenAI-o1, Gemini-2.5-pro) on both in-distribution and out-of-distribution embodied spatial reasoning tasks. Embodied-R also exhibits emergent thinking patterns such as systematic analysis and contextual integration. We further explore research questions including response length, training on VLM, strategies for reward design, and differences in model generalization after SFT (Supervised Fine-Tuning) and RL training.
\end{abstract}

\begin{teaserfigure}
\centering
	\includegraphics[width=0.95 \textwidth]{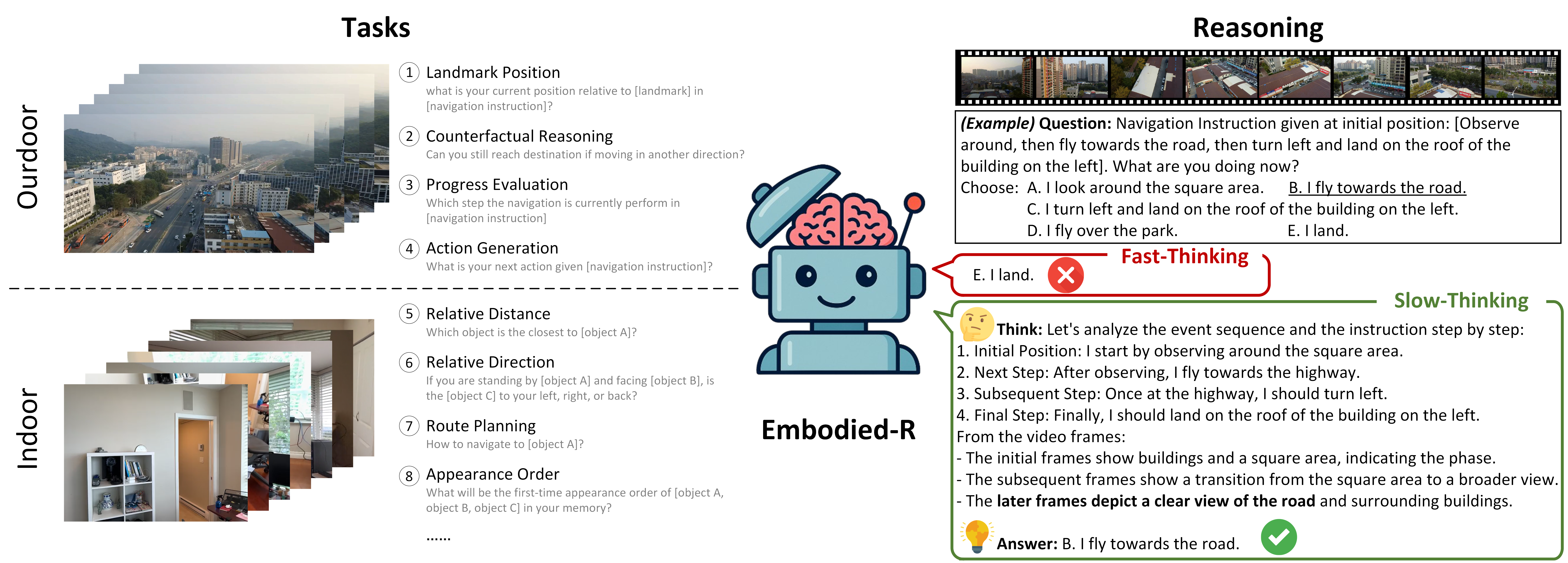}
	\vspace{-3pt}
	\caption{Embodied spatial reasoning: tasks and thinking process. Challenging tasks from public embodied video datasets are identified, encompassing both indoor and outdoor scenarios. We introduce slow-thinking to improve reasoning performance.}
	\label{fig:task}
\end{teaserfigure}

\maketitle
    
\vspace{-0.3cm}
\section{Introduction}
\label{sec:intro}

On the path toward Artificial General Intelligence (AGI)~\cite{fei2022towards}, we hope that pre-trained foundation models can not only perform tasks such as dialogue and image understanding in the cyber world~\cite{achiam2023gpt,team2023gemini} but also develop human-like embodied spatial cognition in the three-dimensional physical world, enabling them to perceive, think, and move~\cite{aubin2022towards,liu2024aligning}. The fundamental way humans achieve spatial cognition is through continuous, dynamic visual observations, akin to video streams~\cite{intriligator2001spatial,liu2023visual}. For example, by observing their surroundings, humans can infer their position relative to nearby objects. Similarly, based on historical visual observations, humans can determine the actions they should take to reach a target destination.

Visual spatial cognition can be divided into two levels: perception and reasoning~\cite{wang2024picture}. Perception refers to ``what is seen", characterized by direct, low-level tasks such as object recognition, edge detection, or color differentiation~\cite{wang2024embodiedscan}. Reasoning, on the other hand, involves ``what is understood" and ``what actions to take", which are indirect and higher-level tasks requiring logical inference and knowledge integration~\cite{zhao2025urbanvideobench}. Examples of reasoning include ``Where did I come from?" (e.g., recalling historical movement trajectories~\cite{mu2023embodiedgpt}), ``Where am I?" (e.g., inferring the spatial relationships between nearby objects and distances~\cite{azuma2022scanqa}), and ``Where do I want to go?" (e.g., planning actions and deciding movements to reach a destination~\cite{chen2024embodied}). While most existing research focuses on improving the perception capabilities of foundation models~\cite{chen2024internvl,bai2025qwen2}, with notable progress, their spatial reasoning abilities remain limited~\cite{chen2024spatialvlm,yang2024thinking}, and methods for enhancement are largely unexplored.

Specifically, video-based spatial reasoning poses several challenges, as follows:  
\begin{itemize}[leftmargin=*]
	\item Reasoning is always built upon perception~\cite{foglia2013embodied,liu2024aligning}. For the studied problem, continuous visual observations impose higher demands on perception. Reasoning cannot be well achieved with faulty perceptions or hallucinations~\cite{wang2024haloquest}. It is challenging to reason when it is already hard to perceive from the videos.
	\item Video data naturally involves complex spatio-temporal relationships, requiring the discovery of object associations across frames and the extraction of semantics relevant to the reasoning task~\cite{fei2024video}. For instance, to navigate to a destination outside the current field of view, one must infer their location from historical visual observations, build a mental map of the environment, develop a high-level plan to determine the direction, and finally decide on specific actions to execute. Existing supervised fine-tuning (SFT) training methods lack supervision for the reasoning process, making it difficult to handle such reasoning tasks~\cite{zhao2025urbanvideobench}.
	\item Embodied visual observations have distinct characteristics. First, understanding disembodied videos, such as movies or TV shows, primarily emphasizes the content within the video, often from a broad and objective perspective~\cite{li2024mvbench}. In contrast, egocentric videos focus on understanding the relationship between the observer and the surrounding environment, often from a constrained first-person perspective~\cite{grauman2022ego4d}. Second, embodied continuous visual observations are generated over time, indicating that embodied perception should rely on sequential inputs rather than aggregating all visual observations for a single input after a prolonged period~\cite{liu2024iof}. Finally, due to the continuity of motion in the physical world, egocentric visual observations also exhibit spatial continuity, meaning there is significant redundancy and repetition between frames. Consequently, directly applying existing multimodal large language models (MLLMs) to embodied videos leads to issues, including loss of generalization and input token limits caused by excessive redundant frames~\cite{abdin2024phi,lin2023video}.
\end{itemize}

Recently, the impressive performance of OpenAI’s o1/o3~\cite{openai2024b} and DeepSeek-R1~\cite{guo2025deepseek} in solving complex reasoning problems(e.g., mathematics, coding, science, etc.) has drawn attention to reinforcement learning (RL) techniques. By incorporating the chain-of-thought (CoT) reasoning process into post-training, large language models (LLMs) demonstrate a "slow-thinking" mode, where they reason thoroughly before generating responses~\cite{team2025kimi,xie2025logic}. Inspired by this, we attempt to introduce ``slow thinking" into embodied video-based spatial reasoning tasks, as shown in Figure \ref{fig:task}.

This brings a new challenge: the trade-off between model size and computational cost. Existing studies suggest a strong correlation between multimodal understanding/perception capabilities and model size~\cite{xu2023mmbench,gao2024embodiedcity,chandrasegaran2024hourvideo}. Since reasoning builds on perception, larger vision-language foundation models should be used as the starting point for training. However, increasing model size leads to often unacceptable computational costs. Additionally, video inputs map to long token sequences, further raising computational demands. Is there a way to leverage the perception capabilities of large-scale models while developing embodied reasoning abilities at a lower computational cost?

Inspired by neuroscience~\cite{zilles2010centenary}, spatial perception and reasoning involve distinct brain regions: visual perception occurs in the visual areas of the occipital lobe~\cite{clarke1990occipital}, basic spatial understanding in the parietal lobe~\cite{fogassi2005parietal}, and complex spatial reasoning in the prefrontal cortex~\cite{donoso2014foundations}. This inspired the design of a collaborative framework with two main components: a large-scale vision-language model (VLM) for perception and a small-scale language model (LM) for reasoning. Based on the continuity of observations, we first propose a key-frame extractor to retain critical information while reducing computational costs. Using a VLM, we sequentially extract semantic information from the frames, which simulates real-world online reasoning while effectively managing the input token length of VLMs for long video inputs. Finally, the semantic information and reasoning question are fed into the small-scale language model, which outputs the reasoning process and final answers.
The small-scale language model is trained with RL, where the reward modeling not only incorporates rule-based rewards inspired by Deepseek-R1-Zero~\cite{guo2025deepseek} but, more importantly, introduces a novel reward for the logical consistency of the reasoning process. In the experiments, we explore seven research questions, covering the framework's performance, RL's role in activating embodied spatial reasoning, and out-of-distribution generalization capabilities.

In general, the main contributions of this paper are as follows:
\begin{itemize}[leftmargin=*]
	\item We propose a \textbf{collaborative} framework for large-scale and small-scale foundation models to address spatial reasoning in the video modality. By decoupling perception and reasoning, the framework leverages the perceptual strength of large-scale foundation models while efficiently enhancing the reasoning capabilities of smaller models in a computationally resource-friendly manner.
	\item This is \textbf{the first work to employ reinforcement learning (RL) to enhance the embodied spatial reasoning abilities of foundation models}. Specifically, we introduce a novel \textbf{logical consistency reward}, which improves the alignment between reasoning processes and generated answers.
	\item Our proposed Embodied-R achieves performance \textbf{comparable to state-of-the-art multimodal large language models (e.g., OpenAI-o1/Gemini-2.5-Pro)} on both in-distribution and out-of-distribution benchmarks. We further investigate \textbf{research questions including the generalization comparison between models trained by SFT \& RL, reward design strategies, etc}.
\end{itemize}

\section{Related Work}

\noindent \textbf{Large Language Model Reasoning.} 
Recently, enhancing reasoning capabilities has become a key focus in large model technologies, demonstrating remarkable performance on tasks such as mathematical and logical problem-solving~\cite{imani2023mathprompter,yang2024qwen2,qwq-32b-preview}. Following the release of OpenAI's o1~\cite{openai2024b}, numerous studies have proposed various technical approaches to achieve similar functionalities, including Chain-of-Thought (CoT)~\cite{wei2022chain}, Monte Carlo Tree Search (MCTS)~\cite{guan2025rstar,zhang2025rest}, distillation~\cite{min2024imitate}, rejection sampling combined with supervised fine-tuning (SFT) or Direct Preference Optimization (DPO)~\cite{setlur2025rl}, among others. Furthermore, Deepseek-r1~\cite{guo2025deepseek} introduced a method to foster the emergence of reasoning abilities in large language models (LLMs) through rule-based rewards combined with reinforcement learning. Similarly, Kimi k1.5~\cite{team2025kimi} proposed a comparable approach, presenting various training techniques, such as curriculum learning. This reinforcement learning paradigm has sparked significant interest, with subsequent works successfully reproducing related results~\cite{xie2025logic,zeng2025simplerl}.

\noindent \textbf{Embodied Spatial Reasoning with VLMs.}
Inspired by the generality of foundation models across various domains~\cite{achiam2023gpt,ahn2024autort}, embodied intelligence aims to develop agents that utilize large multimodal models as their "brains" to achieve perception, navigation, and manipulation in the 3D physical world~\cite{driess2023palm,shah2023lm}. In terms of input, human visual-spatial perception is more akin to continuous RGB observations, similar to video streams~\cite{cheng2024videgothink,suglia2024alanavlm}, rather than static images~\cite{thawakar2025llamav} or point clouds~\cite{wang2024embodiedscan}. Several embodied video benchmarks~\cite{yang2024thinking} demonstrate that, while perception tasks are relatively well-addressed, spatial reasoning tasks—such as spatial relationship inference, navigation, and planning—remain highly challenging. However, existing research~\cite{fei2024video,sun2025video} on video reasoning primarily focuses on disembodied content reasoning, with little emphasis on scenarios involving embodied continuous visual inputs.

\noindent \textbf{Collaboration between large and small models.}
Existing research primarily focuses on addressing the resource consumption and privacy risks associated with large models, as well as the efficiency and performance advantages of small models in specific scenarios~\cite{wang2024comprehensive}. Small models can assist large models in data selection, prompt optimization, and reasoning enhancement~\cite{zhang2023effective,li2024purifying}. The use of small models to detect hallucinations and privacy leakage is explored in~\cite{ulmer2024calibrating,zhao2023automatic}, improving overall system reliability. While our work shares the goal of reducing computational resource demands, it differs by emphasizing the complementary roles of large-scale VLMs in perception and small-scale LMs in enhancing embodied spatial reasoning.

\section{The Embodied-R Method}
\label{sec:method}

\begin{figure*}[h]
	\centering
	\includegraphics[width = \linewidth]{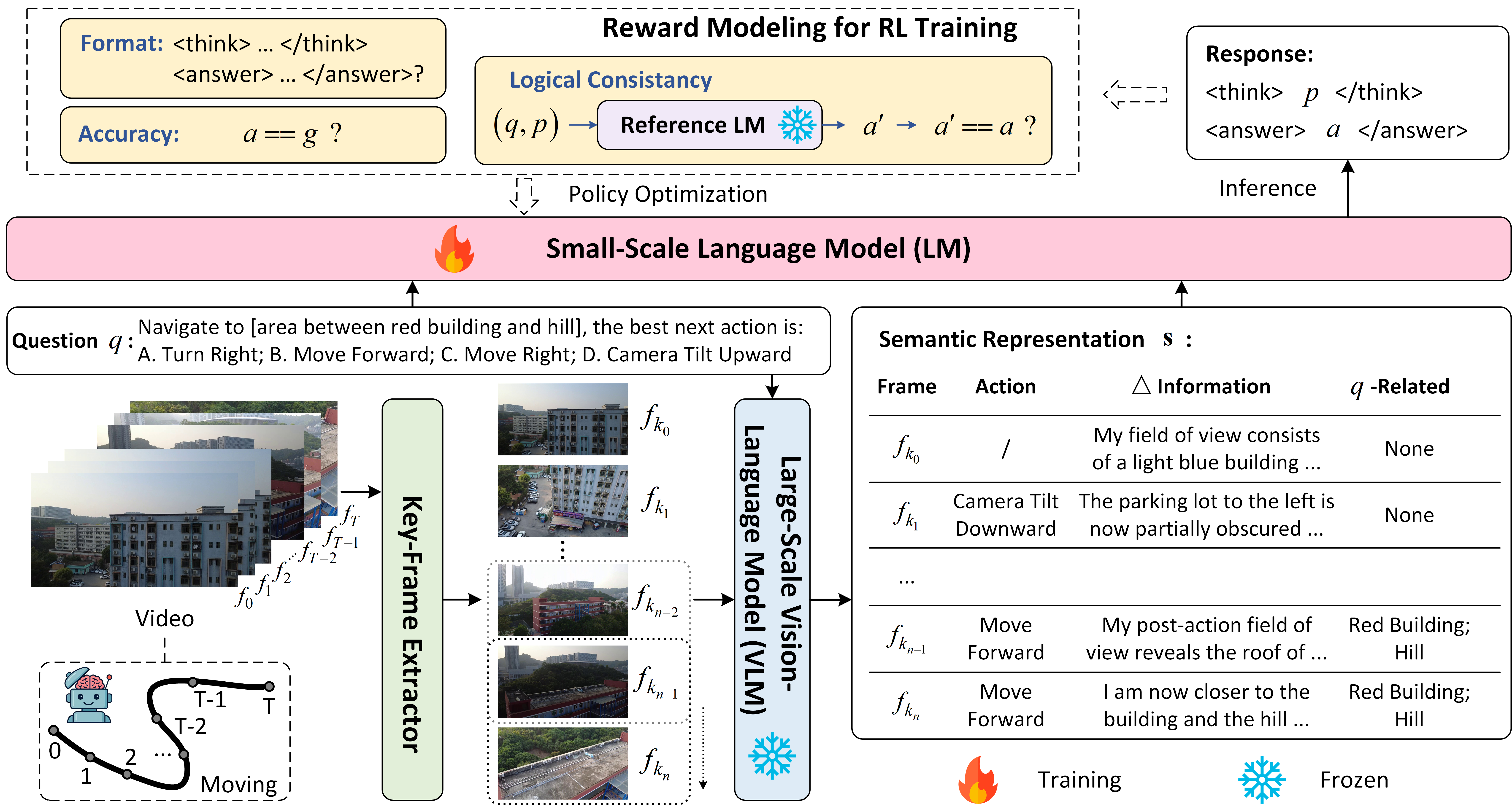}
	\caption{The proposed Embodied-R is a collaborative embodied spatial reasoning framework integrating a Vision-Language Model (VLM) and a Language Model (LM). The separation of perception and reasoning enables us to leverage the perceptual capabilities of large-scale VLMs while training a resource-efficient small-scale LM to activate embodied reasoning through RL. Notably, we introduce a novel logical consistency reward to guide the LM in producing logically coherent reasoning and answer.}
	\label{Fig:framework}
\end{figure*}

We first define the problem of embodied spatial reasoning. Subsequently, we introduce the VLM-based perception module and the LM-based reasoning module. The collaborative framework is shown in Figure \ref{Fig:framework}.

\subsection{Problem Formulation}
In the physical world, an agent moves through space, generating a sequence of video frames (continuous visual observations) $ {\bf{f}} = [f_0, f_1, \dots, f_T]$. Suppose a spatial reasoning problem is denoted as $ q $. Our goal is to build a model that takes $q$ and $\bf{f}$ as inputs and outputs an answer $a$. The answer $a$ is considered correct if it is semantically consistent with the ground truth $g$; otherwise, it is deemed incorrect.

\subsection{Large-Scale VLM-based Perception}

\subsubsection{\textbf{Key-Frame Extractor}}

As the agent moves continuously in space, high sampling frequencies result in significant overlap between consecutive frames. On one hand, the VLM relies on changes in the static objects within the environment across frames to infer the agent's pose variation. On the other hand, excessive overlap between frames leads to increased inference costs for both the VLM and LLM. To address this, we designed a key-frame extractor tailored to the characteristics of embodied videos, selecting key frames that retain overlap while ensuring sufficient information gain between them.

The extraction of key-frames is based on the overlap of visual fields caused by motion continuity. When the agent moves forward, the visual content in the latter frame is expected to overlap with a portion of the former frame, and the reverse is true when moving backward. Similarly, during left or right rotations, the latter frame should partially overlap with the former frame in the horizontal direction, and during upward or downward rotations, the overlap occurs in the vertical direction. Given that the sampling frequency of visual observations is typically much higher than the agent's motion speed, frames generally exhibit significant overlap.

Specifically, a perspective transformation is used to model the geometric relationship between frames. Assuming \( f_t \) is a key-frame, to determine whether \( f_{t+1} \) should also be considered a keyframe, keypoints and descriptors are calculated from \( f_t \) and \( f_{t+1} \) using the Oriented FAST and Rotated BRIEF (ORB) algorithm.
Next, a feature matching algorithm, such as the Brute-Force Matcher, is applied to match the descriptors between the two frames and the Random Sample Consensus (RANSAC) algorithm is employed to estimate the homography matrix. The overlap ratio between two frames is then computed. If overlap ratio is less than a predefined threshold, it indicates significant visual changes between the frames, and \( f_{t+1} \) is marked as a key-frame. Otherwise, the algorithm proceeds to calculate the overlap ratio between \( f_t \) and \( f_{t+2} \). This process continues until a new key-frame is identified, which then becomes the reference for subsequent frames. 
Considering the effect of viewpoint changes, rotations (both horizontal and vertical) result in larger field-of-view variations, leading to more frames being recorded during these movements. If the indices of the extracted keyframes are denoted as \( {\bf{f}}' = \left[ f_{k_0}, f_{k_1}, \dots, f_{k_n} \right] \), the keyframe extraction process can be summarized as:
\begin{equation}
{\bf{f}}' = \text{K-Extractor}(\bf{f}).
\end{equation}

\subsubsection{\textbf{Embodied Semantic Representation}}
Since perceptual capability is positively correlated with model size~\cite{li2024mvbench,zhao2025urbanvideobench,yang2024thinking}, we employ a large-scale VLM to process visual inputs to ensure high-quality perception.
The differential information of each key frame is described sequentially. This approach provides two key benefits: 1) The sequential and dynamic processing aligns better with the characteristics of embodied scenarios, where visual observations are continuously generated over time. At each moment, the model should integrate historical semantic representations with the latest visual observations, rapidly updating the semantic understanding of spatial perception. 2) It facilitates the handling of long videos by avoiding the input token limitations that arise when all frames are processed simultaneously by the VLM. 

Specifically, for the first frame, the VLM identifies the objects present in the scene, their attributes, and their spatial locations. For subsequent frames, both the previous frame and the current frame are input into the VLM to extract key semantic representation ${s_{{k_j}}}$:
\begin{equation}
	{s_{{k_j}}} \sim \psi_\theta (s|{f_{{k_{j - 1}}}},{f_{{k_j}}};q),\:j = 1,2,...,n,
\end{equation}
where ${s_{{k_j}}}$ consists of three items:
\begin{itemize}[leftmargin=*]
	\item \textbf{Action}: Inferring the agent's actions based on the changes in visual observations between consecutive frames.
	\item \textbf{$\triangle \text{Information}$}: Determining changes in the spatial relationships between the agent and known objects, as well as identifying whether new objects appear in the field of view.
	\item \textbf{$q$-related} content: Detecting whether objects or information relevant to the reasoning task appear in the latest field of view.  
\end{itemize}
In this way, we can extract spatial semantic representations \( {\mathbf{s}} = [{s_{{k_0}}},{s_{{k_1}}},...,{s_{{k_n}}}] \) from the keyframe \({\bf{f}}'\).

\subsection{Small-Scale LM-based Reasoning}
Given semantic perception, we can train a training-friendly small-scale language model capable of performing embodied spatial reasoning.
Assuming the small-scale LM is denoted as \( \pi_\theta \), the response \( o \) inferred from the model can be expressed as: $o \sim \pi_\theta(o \mid q, \mathbf{s}).$
	
Our training objective is to ensure that the model adheres to the "think-then-answer" paradigm, where the thinking process is logical, and the answer is correct.
We follow DeepSeek-R1-Zero and adopt a computationally efficient RL training strategy, Group Relative Policy Optimization (GRPO). Besides rule-based format and accuracy rewards, we propose a novel reasoning process reward tailored for embodied reasoning tasks to mitigate reward hacking and enhance the logical consistency between the reasoning process and the final answer. 

\subsubsection{\textbf{Group Relative Policy Optimization}}

For a given query \( q \) and semantic annotation $\mathbf{s}$, GRPO generates a group of outputs \(\{o_1, o_2, \ldots, o_G\}\) using the reference policy \( \pi_{\text{ref}} \). The reference policy typically refers to the original model not trained via GRPO. The policy model \( \pi_\theta \) is then updated by optimizing the following objective:

\begin{equation}
\small
\begin{aligned}
	\mathcal{J}(\theta) & =  \mathbb{E}_{\left( {q,{\bf{s}}} \right) \sim \mathbb{D}, \{o_i\}_{i=1}^G \sim \pi_{\text{old}}(o|{q,{\bf{s}}})} \Bigg[\frac{1}{G}\sum\limits_{i = 1}^G \Bigg( \min \Bigg(\frac{{{\pi _\theta }({o_i}|q,{\bf{s}})}}{{{\pi _{{\rm{old}}}}({o_i}|q,{\bf{s}})}}{A_i}, \\
	&\text{clip} \Bigg( 
	\frac{\pi_\theta(o_i|q,{\bf{s}})}{\pi_{\text{old}}(o_i|q,{\bf{s}})}, 
	1 - \epsilon, 
	1 + \epsilon 
	\Bigg) A_i 
	\Bigg) - \beta \mathcal{D}_{\text{KL}}(\pi_\theta \| \pi_{\text{ref}}) 
	\Bigg) 
	\Bigg], 
\end{aligned}
\end{equation}
where \( \epsilon \) and \( \beta \) are hyperparameters, and \( \mathcal{D}_{\text{KL}}(\pi_\theta \| \pi_{\text{ref}}) \) is KL divergence penalty: $	\mathcal{D}_{\text{KL}}(\pi_\theta \| \pi_{\text{ref}}) = \pi_{\text{ref}}(r_i|q,{\bf{s}}) \log \frac{\pi_{\text{ref}}(r_i|q,{\bf{s}})}{\pi_\theta(r_i|q,{\bf{s}})} - 1$. \( A_i \) represents the advantage corresponding to the output \( o_i \), calculated from the corresponding \( \{r_1, r_2, \ldots, r_G\} \): $	A_i = \frac{r_i - \text{mean}(\{r_1, r_2, \ldots, r_G\})}{\text{std}(\{r_1, r_2, \ldots, r_G\})}$.

\subsubsection{\textbf{Reward Modeling}}
Reward modeling is a critical component of RL algorithms, as their design guides the direction of model optimization. We propose three types of rewards: format reward, accuracy reward, and logical consistency reward. These are designed to respectively guide the model to learn the "think-answer" reasoning pattern, accurate embodied spatial reasoning, and logical consistency between reasoning and the answer.

\noindent \textbf{Format Reward:} 
We aim for the model to output $o_i$ by first producing an embodied reasoning process $p_i$ followed by the final answer $a_i$. The reasoning process and answer are enclosed within <think> </think> and <answer> </answer> tags, respectively:
\begin{tcolorbox}[boxrule=0mm]
	\textit{Please assume the role of an agent. Given a question and a series of frames, you should first think about the reasoning process in the mind and then provide the final answer. The reasoning process and answer are enclosed within <think> </think> and <answer> </answer> tags, respectively, i.e., <think> reasoning process here </think> <answer> answer here </answer>. Ensure that your answer is consistent with and directly derived from your thinking process, maintaining logical coherence between the two sections. The frames represent your egocentric observations from the past to the present. Question: $q$. Video: $\bf{f'}$. Assistant:}
\end{tcolorbox}

A regular expression is applied to evaluate whether $o_i$ meets the specified requirements, thereby generating the format reward ${r'_i}$:
\begin{equation}
	r'_i = 
	\begin{cases} 
		1, & {\rm{if \: format \: is \: correct}}; \\ 
		0, & {\rm{if \: format \: is \: incorrect}}.
	\end{cases}
\end{equation}

\noindent \textbf{Accuracy Reward:}
The accuracy reward ${r''_i}$ model assesses whether the answer $a_i$ is semantically consistent with the ground truth $g$. For example, multiple-choice questions typically have precise and unique answers, which can be easily extracted when the response adheres to the specified format.
\begin{equation}
	r''_i = 
	\begin{cases} 
		1, & {a_i} = g; \\ 
		0, & {a_i} \ne g.
	\end{cases}
\end{equation}

\noindent \textbf{Logical Consistency Reward:}
When using only the format reward and accuracy reward, we consistently observed hacking behaviors. Specifically, for spatial reasoning tasks where the possible answers are limited (e.g., the relative position of an object with respect to the agent's body), cases arise where an incorrect reasoning process $p_i$ leads to a correct answer $a_i$, which is mistakenly assigned a positive reward. As such cases accumulate, the logical consistency of the model's responses deteriorates. To address this issue, we introduce a simple yet effective process reward. Our goal is to ensure a lower bound on logical consistency, such that the reasoning ability of \(\pi_\theta\) should not degrade below that of the reference model \(\pi_{\text{ref}}\). Therefore, when the model's answer is correct ($a_i = g$), we input the question $q$ and reasoning process $p_i$ into the reference model without providing video frames, yielding an answer: 
\begin{equation}
	{a'_i} \sim {\pi _{{\rm{ref}}}}\left( {a|q,{p_i}} \right).
\end{equation}
If ${a'_i}$ is consistent with $a_i$, it indicates that the reasoning process can logically lead to the answer; otherwise, it reflects a logical inconsistency between the reasoning process and the answer.
\begin{equation}
	r'''_i = 
	\begin{cases} 
		1, & a_i = a'_i = g; \\ 
		0, & \text{else}.
	\end{cases}
\end{equation}

\noindent \textbf{Total Reward:}
The total reward is a linear combination of the three rewards mentioned above:
\begin{equation}
	{r_i} = {\omega _1}{r'_i} + {\omega _2}{r''_i} + {\omega _3}{r'''_i}.
\end{equation}

\section{Experiments}
\begin{table*}
	[t]
	\centering
	\caption{Accuracy of Embodied-R and baselines on 8 indoor and outdoor embodied spatial reasoning tasks. The baselines include popular proprietary models, state-of-the-art (SOTA) multimodal reasoning models, open-sourced video-large language models, and models fine-tuned on the same training dataset.}
	\label{tab:acc}
	\setlength{\tabcolsep}{3pt} 
	\begin{tabular}{@{}c@{}} 
		\begin{minipage}[c]{0.58\textwidth} 
\centering \begin{tabular}{r|c|cccccccc}\hline & \multicolumn{1}{l|}{} & \multicolumn{4}{c}{\cellcolor{cyan!20}UrbanVideo-Bench} & \multicolumn{4}{c}{\cellcolor{green!20}VSI-Bench} \\Method &\multicolumn{1}{l|}{Avg.} & \rotatebox{90}{\textit{Landmark Position}} & \rotatebox{90}{\textit{Counterfactual}} & \rotatebox{90}{\textit{Progress Evaluation} } & \rotatebox{90}{\textit{Action Generation}} & \rotatebox{90}{\textit{Relative Distance}} & \rotatebox{90}{\textit{Relative Direction}} & \rotatebox{90}{\textit{Route Planning}} & \rotatebox{90}{\textit{Appearance Order}} \\ \hline Random & 24.0 & 19.7 &25.0 & 21.8 & 16.4 & 25.0 & 36.1 & 28.3 & 25.0 \\ \hline \multicolumn{1}{l|}{\cellcolor[HTML]{F5F5F5}\textit{\textbf{Proprietary Models (API)}}} & \cellcolor[HTML]{F5F5F5} & \multicolumn{8}{c}{\cellcolor[HTML]{F5F5F5}} \\ Qwen-VL-Max[32f] & 34.1 & 44.8 & 49.2 & 38.8 & 29.6 & 28.0 & 33.3 & 29.6 & 28.3\\ GPT-4o[32f] & 35.7 & 36.8 & 44.7 & 34.2 & 33.8 & 37.0 & 41.3 & 31.5 & 28.5\\ Gemini-1.5-Flash[1fps] & 38.3 & 37.8 & 42.4 & 43.3 & 34.4 & 37.7 & 41.0 & 31.5 & 37.8\\ Gemini-1.5-Pro[1fps] & 39.7 & 37.4 & 46.2 & 38.8 & 31.9 & 51.3 & 46.3& 36.0 & 34.6\\ \hline \multicolumn{1}{l|}{\cellcolor[HTML]{F5F5F5}\textit{\textbf{SOTA Reasoning Models (API)}}} & \cellcolor[HTML]{F5F5F5} & \multicolumn{8}{c}{\cellcolor[HTML]{F5F5F5}} \\ OpenAI-o1[32f] & 37.2 & 34.6 & 53.3 & 39.1 & 28.0 & 39.7 & 35.8 & 52.9 & 39.8\\ Gemini-2.5-Pro[1fps] & 40.8 & 40.0 & 75.0 & 38.7 & 23.5 & 42.0 & 34.5 & 52.4 & 63.6 \\ \hline \multicolumn{1}{l|}{\cellcolor[HTML]{F5F5F5}\textit{\textbf{Open-source Models}}} & \cellcolor[HTML]{F5F5F5} & \multicolumn{8}{c}{\cellcolor[HTML]{F5F5F5}} \\ LLaVA-NeXT-Video-7B-hf[32f] & 29.5 & 49.5 & 20.5 & 36.6 & 19.2 & 25.2 & 26.3 & 29.9 & 24.5\\ Phi-3.5-vision-instruct[32f] & 29.0 & 49.2 & 34.8 & 33.2 & 15.6 & 25.4 & 26.5 & 36.9 & 25.2 \\ Kangaroo[64f] & 30.0 & 35.5 & 42.4 & 32.5 & 32.4 & 25.2 & 26.8 & 23.5 & 24.9\\ InternVL2-2B[32] & 24.5 & 19.3 & 45.5 & 29.2 & 20.9 & 25.1 & 25.0 & 32.6 & 23.9\\ InternVL2-8B[32f] & 25.5 &23.1 & 45.5 & 31.5 & 21.4 & 24.7 & 25.7 & 28.3 & 24.8 \\ InternVL2-40B[32f] & 25.8 & 23.2 & 41.7 & 32.4 & 22.3 & 24.9 & 25.7 & 29.4 & 24.5 \\ Qwen2.5-VL-3B-Instruct[1fps] & 33.1 & 32.1 & 47.8 & 34.0 & 31.0 & 27.9 & 32.6 & 39.0 & 38.9 \\ Qwen2.5-VL-7B-Instruct[1fps] & 33.3 & 33.3 & 21.7 & 25.0 & 27.8 & 35.8 & 39.7 & 48.8 & 38.8 \\
Qwen2.5-VL-72B-Instruct[1fps] & 34.9 & 34.7 & 34.8 & 26.4 & 37.7 & 40.8 & 29.0 & 32.5 & 43.9 \\ \hline
\multicolumn{1}{l|}{\cellcolor[HTML]{F5F5F5}\textit{\textbf{Supervised Fine-Tuning}}} & \cellcolor[HTML]{F5F5F5} & \multicolumn{8}{c}{\cellcolor[HTML]{F5F5F5}} \\ Qwen2.5-VL-3B-Instruct[1fps] & 41.7 & 47.7 & 33.4 & 34.8 & 39.2 & 42.6 & 42.3 & 41.2 & 43.9\\ Qwen2.5-VL-7B-Instruct[1fps] & 45.4 & 40.2 & 53.4 & 38.0 & 40.8 & 47.8 & 46.3 & 44.1 & 56.1\\ \multicolumn{1}{l|}{\cellcolor[HTML]{FFE4B5}\textit{\textbf{Proposed Embodied-R}}} & \cellcolor[HTML]{FFE4B5} & \multicolumn{8}{c}{\cellcolor[HTML]{FFE4B5}} \\
VLM-72B + LLM-3B [$\leq$32f] & 51.1 & 55.1 & 59.9 & 39.7 & 47.6 & 50.0 & 44.3 & 36.8 & 72.0\\
\hline\end{tabular}\end{minipage} \hspace{0.7cm} 
		\begin{minipage}[t]{0.38\textwidth} 
\vspace{-6.5cm} \centering \includegraphics[width=\linewidth]{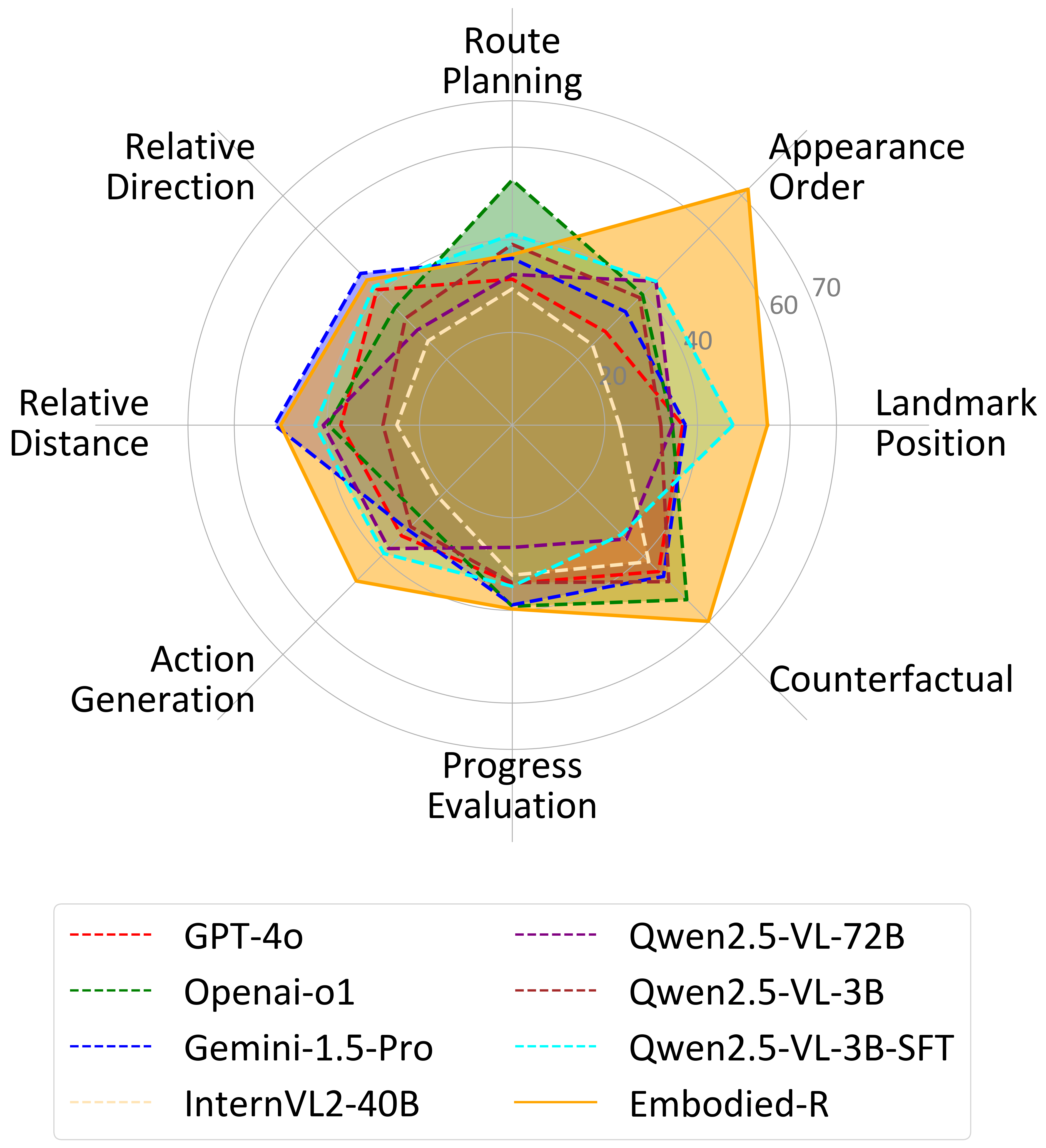} 
\vspace{-0.1cm}

\caption{Ablation of Key-Frame Extractor} \label{tab:ablation_kfe} \vspace{-0.3cm} 
\small
\setlength{\tabcolsep}{3pt} \begin{tabular}{ccccc}\toprule & \makecell{Avg. \\ Frame} & Acc. & \makecell{Training \\ Time} & \makecell{Inference \\ Time} \\ \hline w/o & 32 & 51.1&127.87 h & 243.68 s\\ w & $20.7^{\color{red}\downarrow 11.3}$ & $49.5^{\color{red}\downarrow 1.6}$ &$111.70 h^{\color{red}\downarrow 16.17}$ & $157.55 s^{\color{red}\downarrow 86.13}$ \\ \bottomrule\end{tabular} \vspace{0.47cm} 

\caption{Ablation of Collaboration.} \label{tab:ablation_collaboration} \vspace{-0.17cm} \setlength{\tabcolsep}{1.3pt} \begin{tabular}{cccccccccc}
\hline
     & Avg. & LP   & C    & PE   & AG    & RDist& RDir & RP    & AO\\ \hline
w/o  & 34.8 & 31.8 & 45.7 & 28.3 & 28.1 & 41.0 & 29.7 & 37.5  & 46.0  \\
w    & 51.1 & 55.1 & 59.9 & 39.7 & 47.6 & 50.0 & 44.3 & 36.8 & 72.0\\
$ \bigtriangleup $    & 
\color{green!70!black}+16.3 &  
\color{green!70!black}+23.3 & 
\color{green!70!black}+14.2 & 
\color{green!70!black}+11.4 & 
\color{green!70!black}+19.5 & 
\color{green!70!black}+9.0 & 
\color{green!70!black}+14.6 & 
\color{red}-0.7 & 
\color{green!70!black}+26.0
\\ \bottomrule
\end{tabular}\end{minipage}
	\end{tabular}
\end{table*}

We first provide the details of the experimental setup and then demonstrate the
following: quantitative results, qualitative results, and ablation studies.
These correspond to addressing the following three research questions (RQs):
\begin{itemize}[leftmargin=*]
	\item \textbf{RQ1: How does Embodied-R perform compared to existing video-LLMs?}

	\item \textbf{RQ2: Has Embodied-R learned slow-thinking?}

	\item \textbf{RQ3: What are the contributions of each module?}
\end{itemize}

\begin{figure*}[t!]
	\centering
	\includegraphics[width=0.95\linewidth]{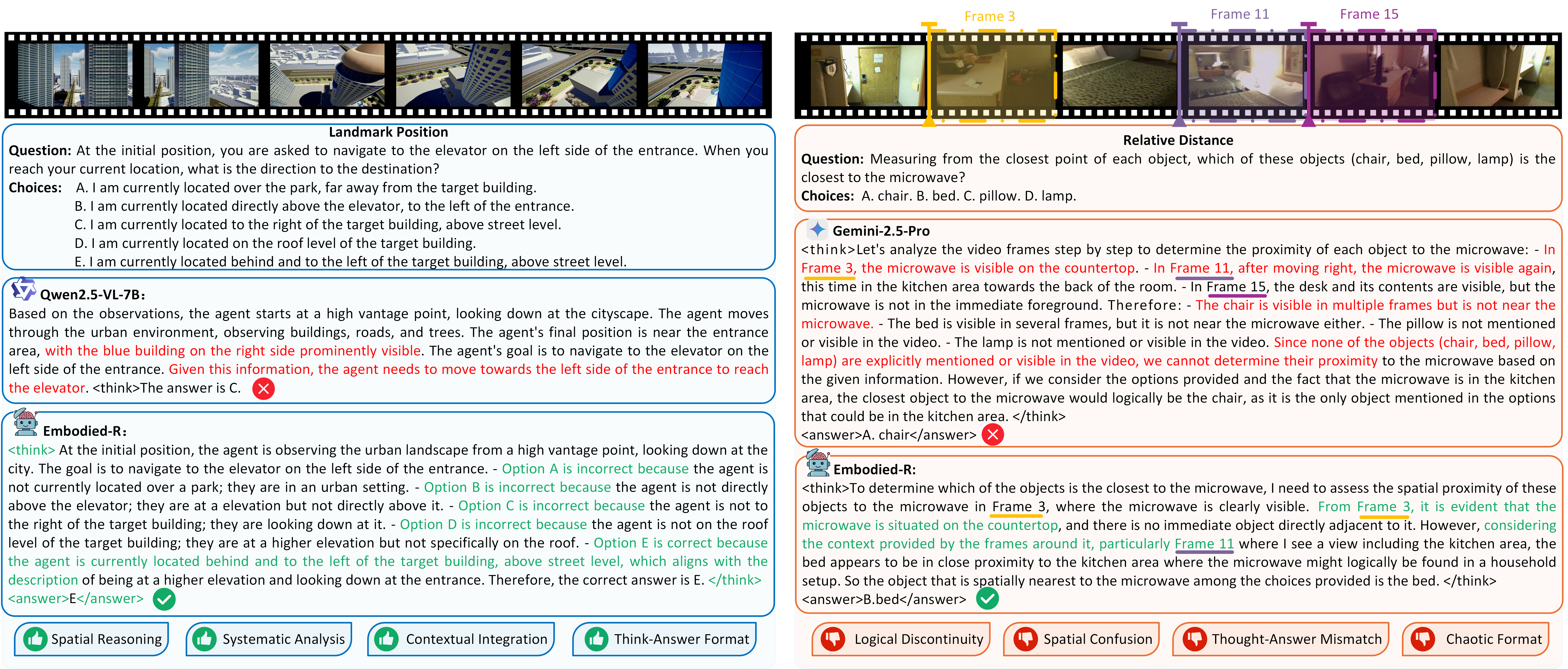}
	\vspace{-10pt}
	\caption{Case Analysis: Embodied-R has initially developed the ability for slow-thinking: it can think before answering, effectively distinguish spatial relationships, provide structured and organized responses, and integrate information across multiple frames for embodied scene analysis.}
	\label{Fig:case}
	\vspace{-5pt}
\end{figure*}

\begin{figure}[t]
	\centering
	\includegraphics[width=3in]{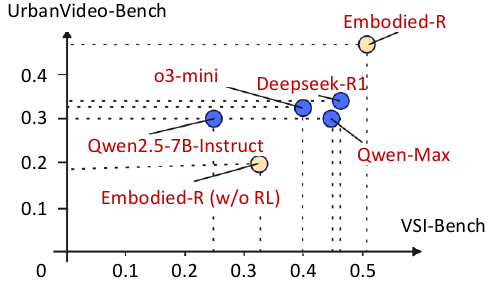}
	\vspace{-5pt}
	\caption{Ablation of RL training and comparison to other language models.}
	\label{Fig:ablation_rl}
	\vspace{-15pt}
\end{figure}

\subsection{Experimental Setup}
\subsubsection{\textbf{Data Preparation}} \label{sec:data_prepare}

We primarily focus on spatial reasoning problems during motion within three-dimensional
physical space to evaluate the effectiveness of our method. For this purpose, we
selected two embodied video datasets as the main training and testing sets:
VSI-Bench~\cite{yang2024thinking},
which contains indoor first-person navigation data.
, and UrbanVideo-Bench~\cite{zhao2025urbanvideobench}, which consists of outdoor embodied
data captured by drones navigating through aerial spaces These datasets provide
diversity in scenarios by incorporating both outdoor and indoor video data. Based
on the content of the tasks, we specifically selected four distinct types of
tasks from each dataset, characterized by long spatial reasoning chains and low accuracy.
These tasks are formulated as multiple-choice question-answering problems, ensuring
determinism in answers to facilitate RL training and allowing direct calculation
of accuracy to evaluate performance. Across eight task categories, the dataset covers
multiple levels of spatial reasoning, comprising a total of 5,415 QA pairs and 1,492
videos. Additionally, we include two out-of-distribution dataset, EgoSchema~\cite{mangalam2023egoschema}
and Egocentric task in MVBench~\cite{li2024mvbench}. EgoSchema is designed for
task-level reasoning from a first-person perspective, with 500 QA pairs and 500 videos
available in its fully open-source portion. MVBench encompasses the embodied
task of egocentric navigation, comprising 200 QA pairs and 200 corresponding videos.
These datasets serves to evaluate the generalization capability of the trained
model.

To ensure comprehensive evaluation, we conducted five repeated experiments. The
dataset was randomly divided into five equal parts and 5-fold cross-validation
is adopted. The final testing results are averaged across the five experiments.
Furthermore, we address the issue of potential semantic bias in the datasets.
For instance, in action generation tasks, forward movement may inherently have a
higher correctness rate than adjusting the gimbal angle, which is a characteristic
of the task itself. To prevent the testing performance from being influenced by
the model learning textual distribution rather than truly understanding the
spatial information in video, we implement an additional filtering step for the testing
set. Specifically, we train a LLM through supervised fine-tuning using only the textual
QA pairs from the training set, without video inputs. If a question in the testing
set can be correctly answered by the fine-tuned LLM but not by the original LLM,
it indicates semantic bias in that QA pair. These biased QA pairs are excluded from
the testing set as they fail to accurately assess the spatial reasoning capabilities
of models.

\subsubsection{\textbf{Implementation Details}}
We use Qwen2.5-3B-Instruct~\cite{yang2024qwen2} as the small-scale LM and Qwen2.5-VL-72B-Instruct~\cite{bai2025qwen2} as large-scale
VLM. Both training and inference processes were conducted using 8 NVIDIA A800-SXM4-40GB
GPUs, with each RL training requiring approximately 90 GPU hours. Other key
hyperparameters for training are as follows: learning rate: 5e-7, temperature:
1.0, train batch size: 32, rollout size: 8, KL coefficient: 0.001, maximum response
length: 2048, input length: 6144. When conducting inference on the test set, the
temperature is set to 0.5.

\subsubsection{\textbf{Three-Stage Training Schedule}}
As for the RL training on the LM, we design a three-stage training schedule to
achieve a smooth improvement in training performance. The primary distinction between
stages lies in the different weight ratios assigned to three types of rewards.

\begin{itemize}[leftmargin=*]
	\item \textbf{Stage 1:} In epochs 1 and 2, the goal is to guide the model to
		follow the "<think> </think> <answer> </answer>" output format. At this
		stage, the weights are set as $\omega_{1}:\omega_{2}:\omega_{3}= 7:3:0$.
		Correct format rewards also assist in locating the answer and reduce
		misjudgment in accuracy. During this phase, the format reward rapidly
		converges to 1.

	\item \textbf{Stage 2:} In epochs 3 and 4, the focus shifts to improving the
		accuracy of the model's responses, guiding the model to produce correct
		reasoning answers. The weights are set as $\omega_{1}:\omega_{2}:\omega_{3}
		= 3:7:0$.

	\item \textbf{Stage 3:} In subsequent 5-12 epochs, the aim is to enhance accuracy
		while simultaneously improving the quality of the "thinking" process,
		ensuring logical consistency between thinking and the answer. The weights
		are set as $\omega_{1}:\omega_{2}:\omega_{3}= 1:7:2$.
\end{itemize}

\subsection{How Does Embodied-R Perform Compared to Existing Video-LLMs?}

To evaluate the effectiveness of the proposed method, in addition to the random baseline, we introduced four categories comprising 17 multimodal large language models capable of processing video inputs:

\begin{itemize}[leftmargin=*]
	\item \textbf{Proprietary Models:} Cost-effective multimodal models with over 100B parameters, including Qwen-VL-Max~\cite{Qwen_Website}, GPT-4o~\cite{OpenAI_API}, Gemini-1.5-Flash~\cite{team2023gemini}, and Gemini-1.5-Pro~\cite{team2023gemini}.
    \item \textbf{SOTA Reasoning Models:} State-of-the-art reasoning models with the highest performance but significant computational cost, including OpenAI-o1~\cite{openai2024b} and Gemini-2.5-Pro~\cite{Gemini_API}.
    \item \textbf{Open-Source Models:} Popular open-source multimodal models, including LLaVA-NeXT-Video-7B-hf~\cite{lin2023video}, Phi-3.5-vision-instruct~\cite{abdin2024phi}, the Internvl2 series~\cite{chen2024internvl}, and the Qwen-VL series~\cite{bai2025qwen2}.
    \item \textbf{Supervised Fine-Tuning (SFT):} Considering the scarcity of embodied video tasks, the aforementioned models may lack exposure to relevant data. Therefore, Qwen2.5-VL-3B-Instruct~\cite{bai2025qwen2} and Qwen2.5-VL-7B-Instruct~\cite{bai2025qwen2} are fine-tuned for these tasks.
\end{itemize}

The results presented in Table~\ref{tab:acc} lead to the following conclusions:

\begin{itemize}[leftmargin=*]
    \item After undergoing RL training on embodied reasoning tasks, our model significantly outperformed proprietary models as well as OpenAI-o1 and Gemini-2.5-Pro by over \textbf{10\%}. Moreover, it consistently demonstrated leading performance across various tasks. These results highlight the considerable \textbf{difficulty of embodied reasoning tasks} and indicate that current reasoning models lack generalization capability for such spatial reasoning challenges. On the other hand, the findings confirm that \textbf{collaborative framework with RL can effectively enhance model reasoning performance in specific domains}, especially for tasks that remain poorly solved.
    
    \item For embodied video reasoning, a highly coupled perception-reasoning problem, the VLM model Qwen2.5-VL-72B-Instruct achieved an accuracy of only 34.9\% through direct inference. In contrast, incorporating a small-scale LM model improved accuracy to 51.1\%. \textbf{Given limited computational resources for training, the collaborative framework proposed in this study provides an effective solution for balancing model size with hardware constraints}.

    \item Under similar computational resource limitations, direct fine-tuning is restricted to models with a size of 7B or smaller. However, the perceptual capacity of small-scale VL models imposes a low upper bound on accuracy compared to Embodied-R. Additionally, fine-tuned models lack the capability for slow-thinking.
\end{itemize}

\subsection{Has Embodied-R Learned Slow-Thinking?}

Beyond the quantitative results, we aim to explore whether spatial reasoning capabilities in the output of Embodied-R are improved. As illustrated in Figure~\ref{Fig:case}, after RL training, Embodied-R demonstrates the following human-like reasoning ways:

\begin{itemize}[leftmargin=*]
    \item \textbf{Spatial Relationship Reasoning}: Accurately inferring the relative spatial relationship between itself and the surrounding environment.  
    \item \textbf{Systematic Analysis}: Breaking down problems into components, presenting answers with a "part-to-whole" structure, and maintaining clear logical organization.

    \item \textbf{Contextual Integration}: Integrating semantic information across different frames to perform comprehensive analysis.

    \item \textbf{Think-Answer Format}: Strictly adhering to a structured process of reasoning before outputting the final answer. 
\end{itemize}
In summary, Embodied-R demonstrates a certain degree of slow-thinking capability in embodied spatial reasoning.

\begin{figure*}[t!]
	\centering
	\includegraphics[width=\linewidth]{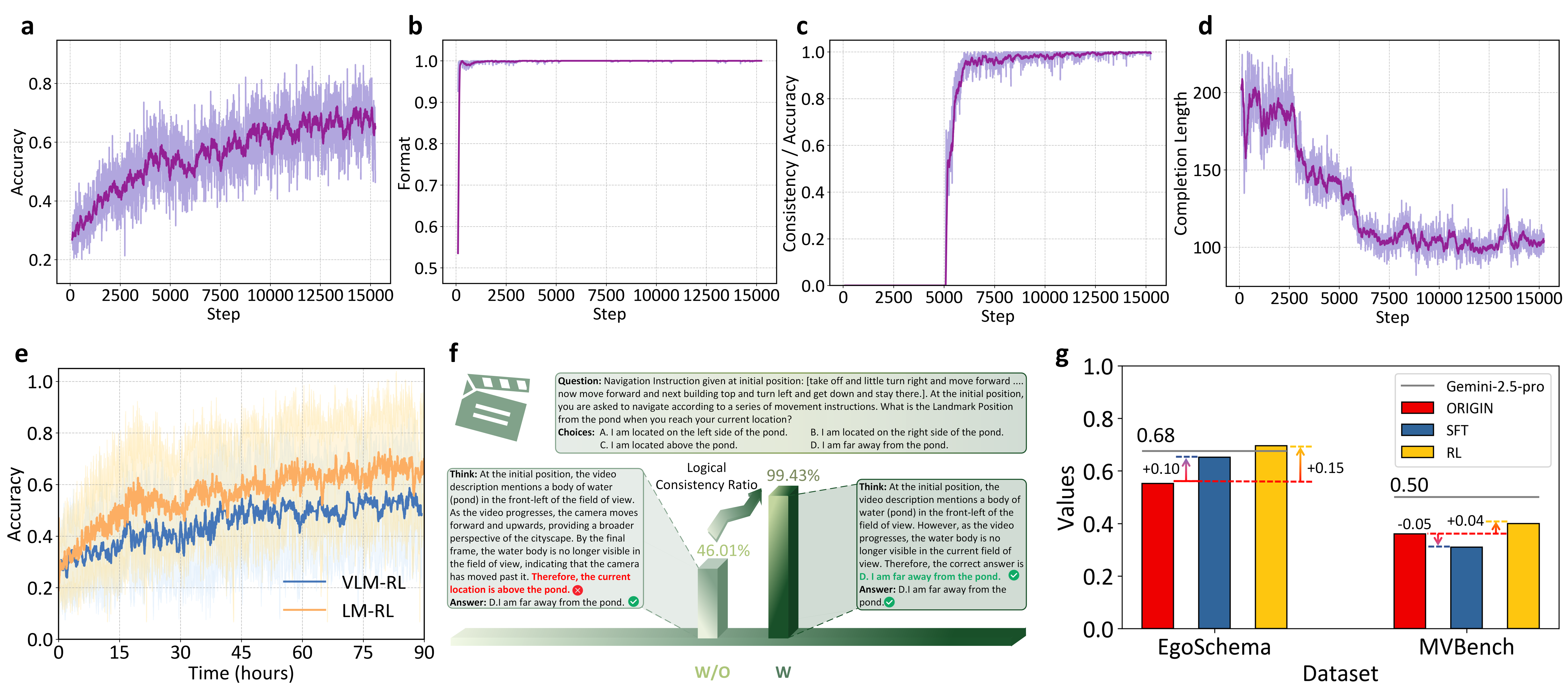}
	\vspace{-15pt}
	\caption{a-d. The GRPO training process (a: accuracy reward; b: format reward; c: ratio of logical consistency reward to accuracy reward; d: response length of validation set).
    e. Comparison of accuracy reward curves for RL training of equivalently sized LM and VLM models.
f. Model performance before and after integrating logical consistency reward.
g. Comparison of generalization performance between models trained with RL and SFT.}
	\label{Fig:further_exploration}
	\vspace{-10pt}
\end{figure*}

\subsection{Contributions of Each Module}
\subsubsection{\textbf{Ablation of Key-Frame Extractor}}

The role of Key-Frame Extractor is to reduce inference time and training time by retaining essential frames and removing redundant ones while maintaining perceptual quality. As shown in Table~\ref{tab:ablation_kfe}, with negligible differences in accuracy, training time is significantly reduced by 8.7\%, and single inference time is reduced by approximately one-third.

\subsubsection{\textbf{Ablation of Collaboration}}

The collaborative framework enables improved reasoning capabilities under limited computational resources for training. With training-free large-scale pre-trained VLMs, it only requires training small-scale LM models to achieve enhanced reasoning performance. As shown in Table~\ref{tab:ablation_collaboration}, with identical key-frame inputs and using the same VLM, Qwen2.5-VL-72B-Instruct, the overall accuracy of collaborative inference is 1.5 times higher than that of the standalone VLM.

\subsubsection{\textbf{Ablation of RL Training}}

RL is central to the LM training in this paper. Without RL training, directly applying the original LM-3B model for reasoning leads to poor performance, as the LM has limited exposure to embodied spatial reasoning data during pretraining. After RL training, the LM achieves significant improvements, with a 27.9\% increase on the UrbanVideo-Bench and a 20.6\% increase on the VSI-Bench benchmarks. 

Given that VLM has already transformed visual inputs into textual representations, we introduced 4 text-based reasoning models (o3-mini~\cite{openai2025o3mini}, Deepseek-R1~\cite{guo2025deepseek}, Qwen-Max~\cite{Qwen_Website}, Qwen2.5-7B-Instruct~\cite{bai2025qwen2}) as baselines to further assess the importance of reasoning capability in the embodied spatial task. The results demonstrate a clear positive correlation between the reasoning ability of the model and its accuracy.
The strong performance of Embodied-R may not only stem from its familiarity with the data distribution but also from its synergy with the representations provided by the VLM. Following training, the small-scale LM becomes more attuned to the VLM-generated representations, which translates into enhanced performance on embodied reasoning tasks.

\vspace{-0.3cm}
\section{Further Exploration}

Building upon the aforementioned experiments, we further explore four intriguing RQs related to embodied video-based RL training:
\begin{itemize}[leftmargin=*]
	\item \textbf{RQ4: What Is the Relationship Between Inference Ability, Aha Moments, and Response Length?}

	\item \textbf{RQ5: Why Not Directly Perform RL Training on VLLMs?}

	\item \textbf{RQ6: Is Accuracy+Format Rewards All You Need?}

    	\item \textbf{RQ7: RL vs SFT when Generalize to Out-of-Distribution (OOD) Embodied Tasks?}
\end{itemize}

\subsection{Relationship Between Inference Ability, Aha Moments, and Response Length?}

The GRPO training process is illustrated in Figure~\ref{Fig:further_exploration}a-d, which correspond to the validation set's accuracy reward, format reward, ratio of logical consistency reward to accuracy reward, and the response length, respectively. Notably, existing pure-text-based reproductions~\cite{xie2025logic,zeng2025simplerl} of DeepSeek-R-Zero models identify inference ability and the "aha moment" as key indicators of emergent reasoning capabilities. However, such phenomena are rarely observed in other multimodal reasoning tasks, such as image-based reasoning~\cite{chen2025r1v,liu2025visual}. This leads us to hypothesize that response length is strongly influenced by the nature of the question itself. For instance, mathematical problems often require multi-step calculations, where increased reasoning length tends to correlate positively with reasoning ability. In contrast, for multimodal reasoning tasks like embodied spatial reasoning, the LM model training process converges toward an optimal range of text output distributions. Concise reasoning patterns may facilitate the embodied spatial reasoning. This highlights the versatility of RL-based post-training method, demonstrating the ability to benefit a wide range of reasoning tasks.

\subsection{Why Not Directly Perform RL on VLLMs?}
We previously attempted direct RL training on the Qwen-VL-3B-Instruct model. As shown in Figure~\ref{Fig:further_exploration}e, under similar training parameters and time, the performance of the VLM was notably inferior to that of the LM. Upon convergence, the VLM achieved an accuracy of 43.8\% on the test set, significantly lower than the LM. The limited perceptual capability of the VLM restricts its potential for reasoning improvements. Therefore, under resource-constrained conditions, collaborative inference integrating models of different scales present a promising solution.

\subsection{Is Accuracy+Format Rewards All You Need?}
According to the Deepseek-R1-Zero, it appears that accuracy and format rewards are enough to guide the model toward correct reasoning. However, during training in our problem, we observed instances of reward hacking, where the model optimizes the answer but the reasoning process leading to that answer is inconsistent with the answer itself. We aim to ensure alignment between the model's reasoning process and its answer, both to enhance generalization and improve the interpretability of the reasoning process. As shown in Figure~\ref{Fig:further_exploration}f, we employ GPT-4o to evaluate the proportion of logically consistent outputs on the test set before and after incorporating a logical consistency reward. This proportion increased from 46.01\% to 99.43\% after the reward was added, demonstrating the value of this approach in addressing embodied spatial multiple-choice reasoning tasks. Moreover, this reward mechanism could potentially be extended to other reasoning tasks prone to answer accuracy hacking during training.

\subsection{RL vs SFT when Generalize to Out-of-Distribution (OOD) Embodied Tasks?}

For small-scale LMs, we aim to explore their generalization performance when trained with SFT instead of RL. To evaluate this, we introduced two OOD datasets: EgoSchema and the egocentric task in MVBench. As discussed in Sections~\ref{sec:data_prepare}, these two OOD datasets differ significantly from the training set in both task content and scene characteristics.
The accuracy results are shown in Figure~\ref{Fig:further_exploration}g. RL-trained models demonstrate generalization ability across both datasets. On the EgoSchema dataset, the RL-trained language model under the Embodied-R framework even achieve performance comparable to the state-of-the-art multimodal reasoning model, Gemini-2.5-Pro. SFT-trained models showed improvement on EgoSchema but a decline on MVBench. This suggests that slow reasoning, as employed in RL models, could be a promising approach to improve the generalization capabilities even for small-scale models.

\section{Conclusion}
To address embodied spatial reasoning tasks, we propose a collaborative framework that leverages the perceptual capabilities of large-scale VLMs and the reasoning potential of compact LMs. Through 90 hours of RL training on a 3B LM using 8 NVIDIA A800-SXM4-40GB GPUs, Embodied-R surpasses OpenAI-o1 by 13.9\% and Gemini-2.5-Pro by 10.3\% on the test set. Other Key findings include: (1) RL training leads to output length convergence, aligning with the requirements of the task; (2) the reasoning upper bound of same-scale VLMs trained with RL is significantly lower than that of Embodied-R, due to inherent limitations in perception; (3) the proposed logical consistency reward enhances reasoning quality; and (4) models trained via RL exhibit stronger generalization on out-of-distribution datasets compared to those trained with SFT.

\bibliographystyle{ACM-Reference-Format}
\bibliography{sample-base}

\clearpage
\appendix
\section{Appendix}
\subsection{Dataset Introduction}
\textbf{UrbanVideo-Bench:}
UrbanVideo-Bench is one of the training and testing datasets designed for embodied reasoning (embodied-r). This benchmark was proposed by Tsinghua University in February 2025. It captures two embodied characteristics of urban environments: complex urban scenes featuring dynamic and static elements, and unique aerial navigation scenarios. The dataset consists of 4 categories and 16 tasks, aimed at evaluating Video-LLMs in terms of recall, perception, reasoning, and navigation capabilities. In our paper, we focus on 4 of these complex tasks for reinforcement learning in video-based learning: \textbf{Landmark Position}, \textbf{Counterfactual Reasoning}, \textbf{Progress Evaluation}, and \textbf{Action Generation}, which represent challenging embodied outdoor tasks.

\textbf{VSI-Bench:}
VSI-Bench is another training and testing dataset for embodied reasoning (embodied-r). Proposed by Fei-Fei Li's team at Stanford in December 2024, this benchmark provides high-quality evaluation metrics for assessing the 3D, video-based, visual-spatial intelligence of multimodal large language models (MLLMs). The dataset comprises 2 categories and 8 tasks designed to evaluate key aspects of spatial reasoning. In our paper, we focus on 4 tasks for reinforcement learning in video-based learning: \textbf{Relative Distance}, \textbf{Relative Direction}, \textbf{Route Planning}, and \textbf{Appearance Order}, all of which are categorized as challenging embodied outdoor tasks.

\textbf{EgoSchema:}
EgoSchema is one of the Out-of-Distribution (OOD) datasets utilized to evaluate the generalization capability of our model. This dataset is specifically designed as a long-form video question-answering benchmark, aimed at assessing modern vision and language systems' ability to understand and reason over extended video content. It provides a rigorous evaluation framework for long video understanding tasks.

\textbf{MVBench:}
MVBench is another Out-of-Distribution (OOD) dataset employed to test the generalization capability of our model. MVBench consists of 20 complex video tasks, offering a comprehensive benchmark for evaluating the video understanding capabilities of existing multimodal models. This dataset is designed to address diverse and challenging scenarios in video-based reasoning.

\subsection{Details of Key-Frame Extractor}

The goal of key-frame extraction is to ensure sufficient information gain between frames while maintaining a certain degree of overlap. The specific process is as follows: 

Step 1: a perspective transformation is used to model the geometric relationship between frames. Assuming \( f_t \) is a key-frame, to determine whether \( f_{t+1} \) should also be considered a keyframe, keypoints and descriptors are calculated from \( f_t \) and \( f_{t+1} \) using the Oriented FAST and Rotated BRIEF (ORB) algorithm:
\begin{equation}
\text{Keypoints}_t, \text{Descriptors}_t = \text{ORB}(f_t),
\end{equation}
\begin{equation}
\text{Keypoints}_{t+1}, \text{Descriptors}_{t+1} = \text{ORB}(f_{t+1}).
\end{equation}
Next, a feature matching algorithm, such as the Brute-Force Matcher, is applied to match the descriptors between the two frames, identifying corresponding keypoint pairs \( \mathbf{l}_t^{\text{key}} \) and \( \mathbf{l}_{t+1}^{\text{key}} \).
Using the matched keypoint pairs, the Random Sample Consensus (RANSAC) algorithm is employed to estimate the homography matrix \( \mathbf{M} \), which maps the content of \( f_{t+1} \) to the coordinate space of \( f_t \).

Step 2: The overlap ratio between two frames is then computed. Assuming the size of each video frame is \( w \times h \), for frames \( f_t \) and \( f_{t+1} \):
\( \mathbf{l}_t = \{[0,0], [w,0], [w,h], [0,h]\} \) represents the four corner points of \( f_t \);
\( \mathbf{l}_{t+1} = \{[0,0], [w,0], [w,h], [0,h]\} \) represents the four corner points of \( f_{t+1} \).
Using the homography matrix \( \mathbf{M} \), the corner points \( \mathbf{l}_{t+1} \) of \( f_{t+1} \) are transformed into the coordinate space of \( f_t \):
\(
\mathbf{l}'_{t+1, i} = \mathbf{M} \cdot \mathbf{l}_{t+1, i}
\),
where \( \mathbf{l}_{t+1, i} = [x, y, 1]^T \) represents the corner points of \( f_{t+1} \) in homogeneous coordinates, and \( \mathbf{l}'_{t+1, i} = [x', y', w']^T \) represents the transformed corner points. The transformed points are further normalized to recover 2D coordinates, resulting in a quadrilateral representing \( f_{t+1} \) in \( f_t \)'s space. In \( f_t \)'s coordinate space, there are two polygons:
Polygon \( L_t \)is defined by the corner points \( \mathbf{l}_t \) of \( f_t \);
Polygon \( L_{t+1}' \)is defined by the transformed corner points \( \mathbf{l}'_{t+1} \).
Thus, the overlap ratio \( c \) is defined as:
\begin{equation}
c = \frac{\text{Area}(L_t \cap L_{t+1}')}{\text{Area}_{\text{total}}}.
\end{equation}

If \( c \) is less than a predefined threshold \(\varepsilon \), it indicates significant visual changes between the frames, and \( f_{t+1} \) is marked as a key-frame. Otherwise, the algorithm proceeds to calculate the overlap ratio between \( f_t \) and \( f_{t+2} \). This process continues until a new key-frame is identified, which then becomes the reference for subsequent frames. 
Considering the effect of viewpoint changes, rotations (both horizontal and vertical) result in larger field-of-view variations, leading to more frames being recorded during these movements. If the indices of the extracted keyframes are denoted as \( {\bf{f}}' = \left[ f_{k_0}, f_{k_1}, \dots, f_{k_n} \right] \), the keyframe extraction process can be summarized as:
\begin{equation}
{\bf{f}}' = \text{K-Extractor}(\bf{f}).
\end{equation}

\subsection{Details of Data Preparation}
\subsubsection{Task Selection Criteria}

In our study, we carefully selected specific tasks that emphasize spatial reasoning capabilities during motion within three-dimensional physical space. The selection process was guided by several key considerations:

\textbf{Focus on Reasoning Processes:} We prioritized tasks that require deep cognitive processing rather than simple recognition or recall. As highlighted in the main text, embodied spatial reasoning involves complex spatio-temporal relationships where agents must discover object associations across frames and extract task-relevant semantics. For instance, navigation tasks require agents to infer their location from historical observations, construct mental maps, develop high-level plans, and determine specific actions—processes that demand sophisticated reasoning capabilities.

\textbf{Diversity in Spatial Contexts:} To ensure comprehensive evaluation, we selected tasks from both indoor (VSI-Bench) and outdoor (UrbanVideo-Bench) environments, providing diverse spatial contexts that test different aspects of embodied reasoning. This diversity is crucial for evaluating the generalizability of our approach across varying spatial scales and environmental complexities.

\textbf{Emphasis on Long Reasoning Chains:} We specifically targeted tasks characterized by long spatial reasoning chains and historically low accuracy rates. These challenging tasks better demonstrate the value of our "slow thinking" approach, which encourages thorough reasoning before generating responses—similar to how recent advances in mathematical and scientific reasoning have benefited from reinforcement learning techniques.

\textbf{Deterministic Evaluation:} All selected tasks were formulated as multiple-choice question-answering problems to ensure determinism in answers, facilitating both RL training and direct calculation of accuracy for performance evaluation.

\subsubsection{Question Filtering Methodology}

To ensure the quality and validity of our dataset, we implemented a rigorous question filtering process:

\textbf{Blind Testing Filter:} We first evaluated questions using an untrained 7B language model without video input (blind selection). Questions that could be correctly answered without visual information were identified as potentially problematic, as they might rely more on textual patterns or common knowledge rather than genuine spatial reasoning based on video content.

\textbf{SFT-based Filtering:} After conducting supervised fine-tuning (SFT) without video inputs, we analyzed which question types showed significant improvement in accuracy. Categories where the model's performance increased substantially without visual information were flagged for removal, as this indicated strong correlations between question text and answers that could be exploited without actual spatial reasoning.

\textbf{Correlation Analysis:} We specifically eliminated question types where:
\begin{itemize}
    \item The model could achieve high accuracy without accessing video content
    \item Performance improved dramatically after text-only SFT training
    \item Question-answer pairs exhibited strong textual patterns that could be exploited without spatial understanding
\end{itemize}

This filtering methodology ensured that our final dataset genuinely tests embodied spatial reasoning capabilities rather than linguistic pattern matching or prior knowledge exploitation. By removing questions with strong text-answer correlations, we created a more challenging and valid benchmark that requires models to truly understand spatial relationships from video content.

\subsection{RL Hyperparameters}
The reinforcement learning (RL) training of Embodied-R requires careful hyperparameter tuning to balance computational efficiency with model performance. We conducted extensive experiments to determine the optimal configuration for our collaborative framework. The key hyperparameters used in our RL training process are summarized in Table~\ref{tab:rl_hyperparams}. These settings were selected to ensure stable training while maximizing the model's embodied reasoning capabilities. Notably, we used a relatively small learning rate (5e-7) to prevent catastrophic forgetting and a moderate KL coefficient (0.001) to maintain proximity to the reference model while allowing sufficient exploration.
\begin{table}[h]
\centering
\caption{Hyperparameters used in reinforcement learning training of Embodied-R.}
\label{tab:rl_hyperparams}
\begin{tabular}{ll}
\toprule
\textbf{Hyperparameter} & \textbf{Value} \\
\midrule
Optimizer & AdamW \\
Learning Rate & 5e-7 \\
Temperature & 1.0 \\
Train Batch Size & 32 \\
Rollout Size & 8 \\
KL Coefficient & 0.001 \\
Maximum Response Length & 2048 \\
Input Length & 6144 \\
Training Epochs & 12 \\
\bottomrule
\end{tabular}
\end{table}

\end{document}